\documentclass[journal]{IEEEtranTIE}
\usepackage{graphicx}
\usepackage{cite}
\usepackage{picinpar}
\usepackage{amsmath}
\usepackage{url}
\usepackage{flushend}
\usepackage[latin1]{inputenc}
\usepackage{colortbl}
\usepackage{soul}
\usepackage{multirow}
\usepackage{pifont}
\usepackage{color}
\usepackage{alltt}
\usepackage[hidelinks]{hyperref}
\usepackage{enumerate}
\usepackage{siunitx}
\usepackage{breakurl}
\usepackage{epstopdf}
\usepackage{pbox}
\usepackage{graphics} % for pdf, bitmapped graphics files
\usepackage{epsfig} % for postscript graphics files
\usepackage{mathptmx} % assumes new font selection scheme installed
\usepackage{times} % assumes new font selection scheme installed
\usepackage{amsmath} % assumes amsmath package installed
\usepackage{amssymb}  % assumes amsmath package installed
\usepackage{tabularx}
\usepackage{bm} 
\usepackage{booktabs}
\usepackage{threeparttable} 

\begin{document}
\title{Information-Theoretic Geometry Optimization and Physics-Aware Learning for Calibration-Free Magnetic Localization}

\author{Wenxuan Xie$^{1}$, Yuelin Zhang$^{1}$, Qingpeng Ding$^{1}$, Jianghua Chen$^{1}$, Jiewen Tan$^{1}$, Jiwei Shan$^{1}$ and Shing Shin Cheng$^{1,*}$% <-this % stops a space
\thanks{Research reported in this work was supported in part by Research Grants Council (RGC) of Hong Kong (CUHK 14217822, CUHK 14207823, CUHK
14211425, and AoE/E-407/24-N) and in part by Innovation and Technology Commission of Hong Kong (MHP/096/22, ITS/235/22, ITS/224/23,
ITS/225/23, and Multi-scale Medical Robotics Center (InnoHK initiative)).}% <-this % stops a space
\thanks{$^{1}$Department of Mechanical and Automation Engineering and T Stone Robotics Institute, The Chinese University of Hong Kong, Hong Kong.}%
\thanks{$^{*}$Corresponding author: Shing Shin Cheng (sscheng@cuhk.edu.hk).}%
}

\maketitle

\begin{abstract}
Wireless localization of permanent magnets enables occlusion-free guidance for medical interventions, yet its practical accuracy is fundamentally limited by two coupled challenges: the poor observability of conventional planar sensor arrays and the simulation-to-reality (Sim-to-Real) gap of learning-based estimators. To address these issues, this article presents a unified framework that combines information-theoretic sensor geometry optimization with physics-aware deep learning. First, a rigorous Fisher Information Matrix (FIM)-based evaluation framework is established to quantify geometry-induced observability limitations. The results show that a staggered split-array topology provides a substantially stronger observability foundation for localization while remaining compatible with practical external deployment. Second, building on this optimized sensing configuration, we propose Phy-GAANet, a calibration-free estimator trained entirely on hardware-aware synthetic data. By incorporating Physics-Informed Features (PIF) for saturation modeling and Geometry-Aware Attention (GAA) for preserving cross-layer vector structure, the network effectively bridges the Sim-to-Real gap. Extensive real-world experiments demonstrate state-of-the-art performance, achieving a position error of 1.84\,mm and an orientation error of 3.18$^{\circ}$ at a refresh rate exceeding 270\,Hz. The proposed method consistently outperforms classical Levenberg--Marquardt solvers and generic convolutional baselines, particularly in suppressing catastrophic outliers and maintaining robustness in challenging near-field boundary regions. Beyond the proposed network, the FIM-guided analysis also provides a framework for sensor geometry design in magnetic localization systems under practical deployment constraints.
\end{abstract}


\markboth{IEEE TRANSACTIONS ON INDUSTRIAL ELECTRONICS}%
{}

\definecolor{limegreen}{rgb}{0.2, 0.8, 0.2}
\definecolor{forestgreen}{rgb}{0.13, 0.55, 0.13}
\definecolor{greenhtml}{rgb}{0.0, 0.5, 0.0}

\section{Introduction}
\label{sec:introduction}

\IEEEPARstart{W}{ireless} tracking of permanent magnets has emerged as a pivotal enabling technology for medical robotics and healthcare due to its ability to provide real-time, precise localization in an inherently occlusion- and radiation-free manner. This capability has enabled applications ranging from monitoring muscle dynamics~\cite{doi:10.1126/scirobotics.abg0656}, to localizing magnetic catheter tips~\cite{11206571}, and navigating wireless capsule endoscopes (WCEs) for gastrointestinal examinations~\cite{than2012review}. Despite their diverse operational environments, the core functionality of these systems relies on establishing a reliable mapping between magnetic field measurements and the magnet's spatial pose.

In the realm of wireless localization, the magnetic dipole model formulates tracking as an inverse problem, typically solved using the Levenberg--Marquardt (LM) algorithm because of its balance between convergence speed and precision~\cite{hu2010cubic,taylor2019low,su2017investigation}. However, as a local optimizer, LM is inherently dependent on initialization quality, making it prone to stagnation in local minima under environmental noise or interference~\cite{11119162}. To improve convergence robustness, hybrid strategies integrating metaheuristics (e.g., Particle Swarm Optimization (PSO), Whale Optimization Algorithm (WOA)) or neural-network-based initialization have been explored~\cite{MIRJALILI201651,qin2022hffnn,lv2021improving}. Nevertheless, these paradigms impose a prohibitive trade-off: global search mechanisms incur computational latency that is ill-suited for real-time control while still failing to guarantee optimality, with failure rates reaching 14.9\% in complex scenarios~\cite{lv2021improving}. Consequently, despite analytical accelerations~\cite{taylor2019low,9639620}, the dependence on robust initialization remains an unresolved bottleneck in optimization-based frameworks.

To overcome the initialization sensitivity and iterative latency of classical optimization, recent studies have increasingly turned to deep learning, establishing end-to-end frameworks capable of inverting the magnetic model in a single inference pass~\cite{su2023amagposenet, xie2025theoretical, 11246085, 11015676, 11151691, 11206571, sebkhi2019deep, 11358394}. This evolution---from simple regressions to more sophisticated architectures integrating coordinate attention~\cite{11015676} or symmetry-aware convolutions~\cite{11151691}---has improved spatial feature extraction. However, the impracticality of acquiring large-scale clinical datasets~\cite{sebkhi2019deep} necessitates training on synthetic data, inevitably introducing a simulation-to-reality (Sim-to-Real) gap. Recent efforts have therefore incorporated physical priors directly into network design. Representative strategies include physics-informed residual networks (PIRN) for model correction~\cite{11246085}, few-shot transfer learning~\cite{11206571, su2023amagposenet}, and noise injection for calibration-free localization~\cite{xie2025theoretical}.

Despite recent algorithmic advances, localization accuracy remains fundamentally constrained by the physical measurement setup and often degrades sharply as vertical distance increases~\cite{su2017investigation, xie2025theoretical}. This highlights the central role of sensor geometry in determining system observability. Prior studies have explored planar placement optimization~\cite{7815347}, reconfigurable curvature-conforming arrays~\cite{10643774, 10643424}, and fully enclosed cubic arrays~\cite{hu2010cubic}. However, planar analyses neglect the observability loss induced by vertical displacement, reconfigurable arrays introduce additional sensing and modeling complexity, and enclosed cubic designs interfere with external magnetic actuation. More importantly, most existing studies focus on idealized workspaces where the dipole model remains reliable, leaving limited understanding of how three-dimensional spatial diversity can improve localization when the target approaches workspace boundaries and the underlying physical model begins to deteriorate.

At the same time, geometry optimization alone is insufficient for practical deployment. Theoretical observability gains are often weakened by unavoidable hardware imperfections, such as sensor manufacturing tolerances and installation errors, which degrade data quality and widen the Sim-to-Real gap. Although calibration methods can partially compensate for these effects~\cite{su2017investigation, hu2006calibration}, they typically rely on system-specific tuning and therefore remain difficult to scale for widespread clinical adoption~\cite{xie2025theoretical}. Consequently, a practical solution must jointly address geometric observability and real-world robustness.

To bridge the gap between theoretical observability and practical utility, we present a comprehensive framework that synergizes information-theoretic geometric optimization with physics-guided deep learning. By jointly addressing geometric degradation and the Sim-to-Real gap, the proposed framework enables robust calibration-free localization. The main contributions are summarized as follows:

\begin{enumerate}
    \item A rigorous evaluation framework utilizing the Fisher Information Matrix (FIM) is established to quantitatively screen sensor topologies. This analysis identifies the inherent information deficiency of conventional planar arrays and demonstrates the effectiveness of a novel staggered split-array topology. The optimized configuration increases the intrinsic information density by an order of magnitude and significantly lowers the theoretical lower bound of tracking error, thereby providing a superior physical foundation for high-precision localization.

    \item A novel physics-aware learning architecture, Phy-GAANet, is proposed to asymptotically approach the theoretical precision limits enabled by the optimized geometry. By integrating Physics-Informed Features and Geometry-Aware Attention to bridge the Sim-to-Real gap, the network achieves state-of-the-art accuracy without requiring real-world data collection or hardware calibration.
\end{enumerate}

\section{Method}
\label{sec:Method}

\begin{figure}[t]
\centering
\includegraphics[width=2.8in]{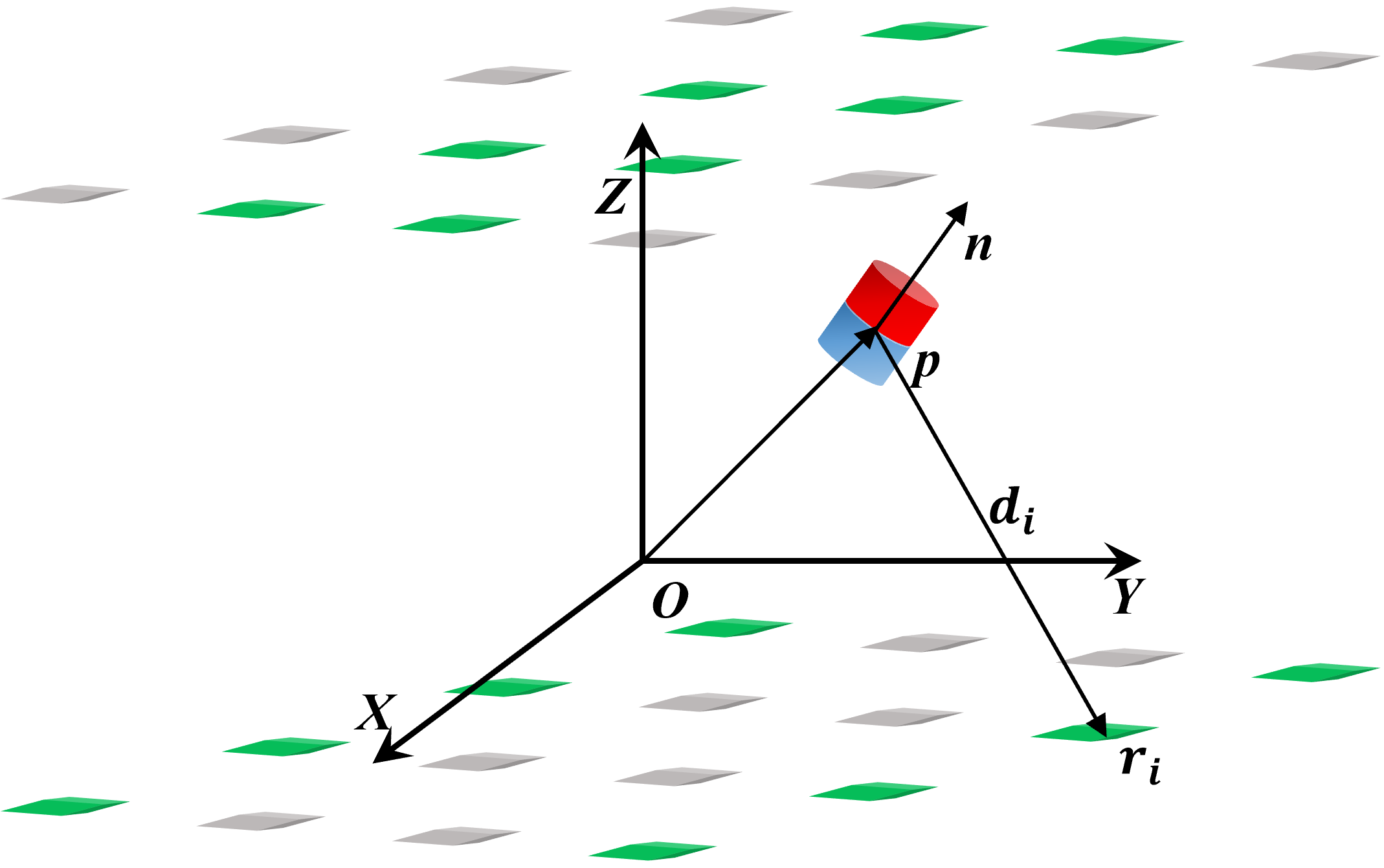}
\vspace{-3mm}
\caption{Schematic representation of the 5-DOF localization system based on the magnetic dipole model. The state of the permanent magnet is defined by its center position $\mathbf{p}$ and orientation vector $\mathbf{n}$. The displacement vector $\mathbf{d}_i$ originates from the magnet towards the $i$-th sensor located at $\mathbf{r}_i$ in the dual-layer sensor array. The staggered height configuration of the sensors (indicated by green and grey colors) is designed to improve the gradient information in the vertical direction.}
\label{fig_1}
\vspace{-4mm}
\end{figure}

\begin{figure*}[t]
\centering
\includegraphics[width=0.95\textwidth,keepaspectratio]{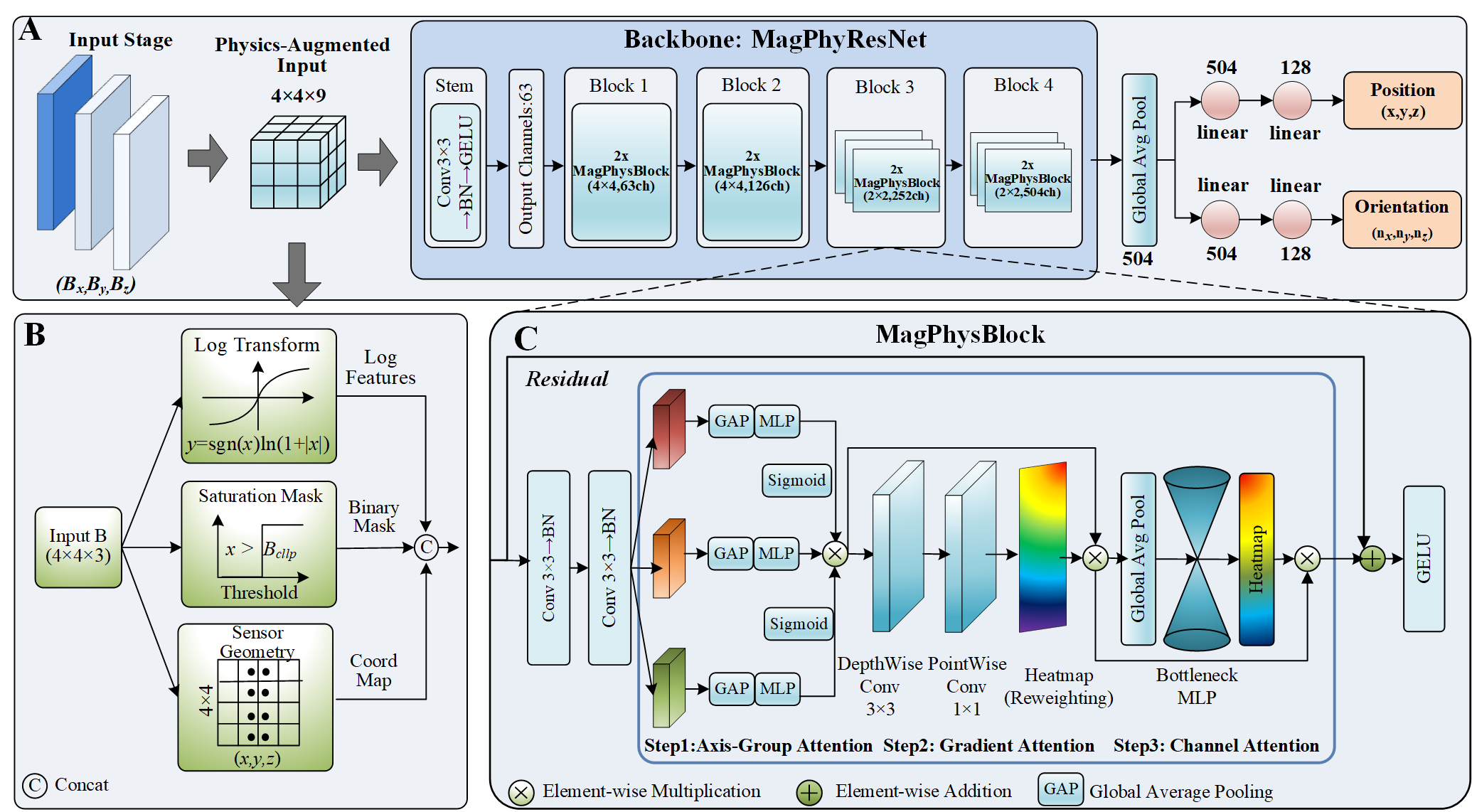}
\caption{Overall architecture of the proposed Phy-GAANet.
\textbf{(A)} The overall network pipeline. The model accepts a physics-augmented input tensor ($4 \times 4 \times 9$), processes it through the MagPhyResNet backbone, and outputs position ($\mathbf{p}$) and orientation ($\mathbf{n}$) estimates via decoupled MLP heads.
\textbf{(B)} Physics-augmented input stage. To handle the unique characteristics of magnetic signals, three feature maps are concatenated: Log Transform (for compressing the high dynamic range), Saturation Mask (indicating clipped sensor readings), and Sensor Geometry (providing spatial coordinates).
\textbf{(C)} Detailed structure of the MagPhysBlock. This residual building block integrates a three-stage attention mechanism to recalibrate features: Step 1 groups axes to suppress directional noise (Axis-Group Attention); Step 2 employs depth-wise convolutions to capture spatial field variations (Gradient Attention); and Step 3 applies a Squeeze-and-Excitation (SE) block for global feature weighting (Channel Attention). Symbols $\otimes$ and $\oplus$ denote element-wise multiplication and element-wise addition, respectively.}
\label{fig_2}
\vspace{-4mm}
\end{figure*}

\subsection{Magnetic Dipole Field Modeling}
\label{ssec:Dipole}

As illustrated in Fig.~\ref{fig_1}, we consider the localization of a cylindrical permanent magnet using a stationary sensor array. The external magnetic field is approximated using the magnetic dipole model, which remains accurate when the magnet-to-sensor distance exceeds the magnet's characteristic dimensions (typically more than eight times the radius~\cite{Hu2006}).

The field intensity is governed by the magnet's physical properties, specifically radius $r$, length $l$, surface magnetization $M$, and the permeability of the surrounding medium. These intrinsic parameters are encapsulated in the magnetic strength coefficient, defined as $B_T = (\mu_r \mu_0 \pi r^2 l M)/4\pi$, where $\mu_0$ and $\mu_r$ denote the vacuum permeability and relative permeability, respectively. Accordingly, the magnetic flux density $\mathbf{B}_i \in \mathbb{R}^3$ measured at the $i$-th sensor position $\mathbf{r}_i$ is expressed as
\begin{equation}
\label{eq:B_vector}
    \mathbf{B}_{i}(\mathbf{p}, \mathbf{n}) = B_T
    \left(
        \frac{3 (\mathbf{n} \cdot \mathbf{d}_{i}) \mathbf{d}_{i}}{\|\mathbf{d}_{i}\|^5}
        - \frac{\mathbf{n}}{\|\mathbf{d}_{i}\|^3}
    \right),
\end{equation}
where $\mathbf{d}_{i} = \mathbf{r}_i - \mathbf{p}$ denotes the displacement vector from the magnet center to the $i$-th sensor, and $\mathbf{n}$ is the unit orientation vector ($\|\mathbf{n}\|_2 = 1$).

To facilitate the subsequent gradient-based observability analysis, we adopt a minimal parameterization of the orientation $\mathbf{n}$ using yaw ($\psi \in [0, 2\pi)$) and pitch ($\theta \in [0, \pi]$):
\begin{equation}
\label{eq:n_vector_parametrization}
\mathbf{n}(\psi, \theta) =
\begin{bmatrix}
    \cos\psi \sin\theta, & \sin\psi \sin\theta, & \cos\theta
\end{bmatrix}^{\top}.
\end{equation}
This formulation yields the minimal state vector $\mathbf{x} = [\mathbf{p}^{\top}, \psi, \theta]^{\top} \in \mathbb{R}^{5}$, which inherently satisfies the unit-norm constraint and simplifies the subsequent optimization and observability analysis.

\subsection{Geometry-Induced Observability Analysis}
\label{ssec:geom_observability_method}

The localization fidelity of magnetic dipole tracking is fundamentally governed by the spatial configuration of the sensor array. Conventional planar arrays confine sensors to a single plane, inherently limiting the observability of magnetic field gradient variations along the normal axis ($Z$). This geometric constraint leads to an ill-conditioned FIM and, in practice, to a significant attenuation of observable information along the normal direction. While volumetric configurations, such as cubic arrays~\cite{hu2010cubic}, can significantly improve localization accuracy, they inevitably obstruct the workspace and therefore hinder external magnetic actuation.

To address this trade-off, we propose a ``minimal intrusion'' design strategy, namely the staggered split-array configuration. By strategically alternating sensor elevations between two distinct levels ($Z_{\text{BOT}}$ and $Z_{\text{TOP}}$), this topology enhances the observable information content without obstructing the lateral workspace required for clinical manipulation. To evaluate this design quantitatively and to establish the theoretical precision limits of candidate geometries, we employ an analysis framework based on the Cram\'{e}r--Rao Lower Bound (CRLB).

Given the minimal state vector $\mathbf{x} = [\mathbf{p}^\top, \psi, \theta]^\top \in \mathbb{R}^{5}$ defined in Section~\ref{ssec:Dipole}, and an array comprising $N$ triaxial magnetometers, the complete observation vector is denoted as $\mathbf{y}(\mathbf{x}) = [\mathbf{B}_1^\top, \dots, \mathbf{B}_N^\top]^\top \in \mathbb{R}^{3N}$. Assuming that the measurements are corrupted by independent and identically distributed (i.i.d.) additive white Gaussian noise with standard deviation $\sigma$, the FIM is given by
\begin{equation}
\label{eq:FIM_definition}
    \mathbf{F}(\mathbf{x}) = \frac{1}{\sigma^2}\mathbf{J}(\mathbf{x})^\top \mathbf{J}(\mathbf{x}),
\end{equation}
where $\mathbf{J}(\mathbf{x}) \in \mathbb{R}^{3N \times 5}$ is the Jacobian matrix containing the partial derivatives of the magnetic field model with respect to the state variables, i.e., $\partial \mathbf{B} / \partial \mathbf{x}$. This matrix captures the local sensitivity of the sensor array to state perturbations.

According to the CRLB, the error covariance of any unbiased estimator $\widehat{\mathbf{x}}$ is lower bounded by the inverse FIM, i.e., $\mathrm{Cov}(\widehat{\mathbf{x}}) \succeq \mathbf{F}^{-1}(\mathbf{x})$. To rigorously benchmark candidate geometries, we define four scalar metrics that characterize both the estimation bound and the information geometry of the sensing system:
\begin{itemize}
    \item \textbf{Position RMSE Bound}: Computed as $\sqrt{\mathrm{Tr}(\mathbf{F}^{-1}_{1:3,1:3})}$, representing the theoretical lower bound of the position root-mean-square error (RMSE).
    \item \textbf{Orientation RMSE Bound}: Computed as $\sqrt{\mathrm{Tr}(\mathbf{F}^{-1}_{4:5,4:5})}$, representing the theoretical lower bound of the orientation RMSE (converted to degrees).
    \item \textbf{Minimum Eigenvalue ($\lambda_{\min}$)}: Characterizes the information density in the least observable direction, i.e., the worst-case sensitivity. A vanishing $\lambda_{\min}$ indicates geometric degeneracy.
    \item \textbf{Condition Number ($\kappa$)}: Defined as $\lambda_{\max}/\lambda_{\min}$, quantifying the numerical stability and isotropy of the sensing system.
\end{itemize}
These metrics constitute a theoretical screening mechanism. By evaluating them \textit{in silico}, we identify geometries with superior information density and more favorable precision bounds, thereby ensuring that the subsequent hardware integration and neural network training are grounded on an optimized geometric foundation.

\subsection{Network Architecture: Phy-GAANet}
\label{sec:method_arch}

To address real-time magnetic localization under sensor noise and calibration errors, we propose the Physics-Guided Geometry-Aware Attention Network (Phy-GAANet). As illustrated in Fig.~\ref{fig_2}, the architecture is designed to learn robust magnetic field representations directly from raw, uncalibrated sensor array data. Unlike generic convolutional neural networks (CNNs), Phy-GAANet incorporates domain-specific physical priors---specifically magnetic-field decay and sensor-array geometry---into both the input representation and the attention mechanisms. The network consists of three key stages: a physics-augmented input, a residual backbone enhanced with magnetic gradient attention, and decoupled regression heads.

\begin{table}[t]
\caption{Detailed Architecture Configuration of Phy-GAANet}
\label{tab:arch_details}
\centering
\begin{tabular}{c|c|c|c}
\hline
\textbf{Stage} & \textbf{Output Size} & \textbf{Layer Configuration} & \textbf{Stride} \\
\hline
Input & $4 \times 4 \times 9$ & $[\mathbf{X}_{log}; \mathbf{X}_{sat}; \mathbf{X}_{geo}]$ & - \\
\hline
Stem & $4 \times 4 \times 63$ & Conv$3\times3$, BN, GELU & 1 \\
\hline
Stage 1 & $4 \times 4 \times 63$ & $\begin{bmatrix} \text{MagPhysBlock} \times 2 \end{bmatrix}$ & 1 \\
\hline
Stage 2 & $4 \times 4 \times 126$ & $\begin{bmatrix} \text{MagPhysBlock} \times 2 \end{bmatrix}$ & 1 \\
\hline
Stage 3 & $2 \times 2 \times 252$ & $\begin{bmatrix} \text{MagPhysBlock} \times 2 \end{bmatrix}$ & 2 \\
\hline
Stage 4 & $2 \times 2 \times 504$ & $\begin{bmatrix} \text{MagPhysBlock} \times 2 \end{bmatrix}$ & 1 \\
\hline
Heads & $1 \times 1 \times 3$ & $\begin{matrix} \text{Pos: MLP}(504 \to 128 \to 3) \\ \text{Ori: MLP}(504 \to 128 \to 3) \end{matrix}$ & - \\
\hline
\end{tabular}
\vspace{-4mm}
\end{table}

To handle the high dynamic range of magnetic flux density ($B \propto r^{-3}$) and hardware saturation, we construct a multi-channel input tensor $\mathbf{X}_{in} \in \mathbb{R}^{H \times W \times C_{in}}$, where $H=W=4$ and $C_{in}=9$, comprising three physical components. First, to compress the dynamic range and enhance sensitivity to weak signals, we apply a signed logarithmic transformation to the raw magnetometer readings $\mathbf{B}_{raw}$, denoted as $\mathbf{X}_{log} = \mathrm{sgn}(\mathbf{B}_{raw}) \cdot \ln(1 + |\mathbf{B}_{raw}|)$. Simultaneously, we address sensor saturation by introducing a binary mask $\mathbf{X}_{sat} = \mathbb{I}(|\mathbf{B}_{raw}| \geq B_{clip})$, where $B_{clip}$ is set to $1900\,\mu\mathrm{T}$, explicitly informing the network of invalid or clipped readings. Furthermore, the physical coordinates $(p_x, p_y, p_z)$ of each sensor are embedded as $\mathbf{X}_{geo}$ to provide an absolute spatial reference frame. These components are concatenated and projected into a latent feature space via a ``Physics Stem'' consisting of a $3\times3$ convolution, Batch Normalization (BN), and Gaussian Error Linear Unit (GELU) activation.

A key component of the backbone is the MagPhysBlock module (detailed in Fig.~\ref{fig_2}), which replaces standard spatial attention mechanisms. Inspired by analytical dipole models that rely on magnetic gradients and vector components, MagPhysBlock utilizes its internal Geometry-Aware Attention to sequentially recalibrate features across three dimensions. It begins by grouping feature channels into orthogonal triplets $(x,y,z)$ and reweighting them via an axis-specific Multi-Layer Perceptron (MLP) to suppress noise-dominated axes. This is followed by a gradient-aware spatial attention mechanism that employs a $3\times3$ depth-wise convolution to aggregate neighboring information and highlight regions with steep field gradients. Finally, a Squeeze-and-Excitation (SE) block recalibrates channel weights. Formally, for an input feature map $\mathbf{F}$, the recalibrated feature map is written as
\begin{equation}
    \mathbf{F}' = \mathbf{F} \otimes \mathcal{M}_{axis}(\mathbf{F}) \otimes \mathcal{M}_{grad}(\mathbf{F}) \otimes \mathcal{M}_{channel}(\mathbf{F}),
\end{equation}
where $\otimes$ denotes element-wise multiplication and $\mathcal{M}_{(\cdot)}$ denotes the corresponding attention map.

For feature extraction, we employ a modified ResNet-18 architecture tailored to the low spatial resolution ($4\times4$) of the sensor array, as detailed in Table~\ref{tab:arch_details}. The network comprises four residual stages, with downsampling strategically delayed until the third stage to preserve spatial information. The network ends with two decoupled task-specific heads to prevent negative transfer: one for position estimation, which relies primarily on field magnitude and gradients, and one for orientation estimation, which depends on vector directionality. Each head consists of a global average pooling layer followed by a two-layer MLP (hidden dimension: 128) with GELU activation, outputting the position vector $\mathbf{p} \in \mathbb{R}^{3}$ and the orientation unit vector $\mathbf{n} \in \mathbb{R}^{3}$, respectively.

\subsection{Loss Function}

To jointly optimize position and orientation estimation, we employ a weighted multi-task objective. Let $\hat{\mathbf{p}}, \hat{\mathbf{n}}$ denote the predicted position and orientation, and let $\mathbf{p}, \mathbf{n}$ denote the corresponding ground truth. The total loss is defined as
\begin{equation}
    \mathcal{L}_{total} = \lambda_{pos} \|\hat{\mathbf{p}} - \mathbf{p}\|_1 + \lambda_{ori} \|\hat{\mathbf{n}} - \mathbf{n}\|_1,
\end{equation}
where $\|\cdot\|_1$ denotes the $L_1$ norm. We empirically set the balancing weights to $\lambda_{pos}=1000$ and $\lambda_{ori}=10$ to balance the gradient scales of the two tasks, since position is measured in meters and typically has a much smaller numerical range than the normalized orientation vector.

\section{EXPERIMENTS}
\label{Sec:Experiments}

\subsection{Geometry-Induced Observability Evaluation}
\label{ssec:geom_observability_results}

To validate the theoretical framework established in Section~\ref{ssec:geom_observability_method} and to quantify the impact of sensor topology on localization limits, we performed a rigorous \textit{in silico} evaluation. The primary objective was to determine whether a ``minimal intrusion'' strategy, i.e., redistributing sensors solely along the vertical axis, provides sufficient geometric diversity to improve observability along the normal direction. We benchmarked three representative geometries ($N=16$) within the target clinical workspace defined by $x, y \in [-50, 50]$\,mm and $z \in [50, 150]$\,mm. The evaluation used 200,000 samples generated via Latin Hypercube Sampling (LHS) under a realistic noise floor of $\sigma = 10\,\mu\text{T}$.

All configurations share an identical $4 \times 4$ grid footprint uniformly distributed over a $100 \times 100$\,mm$^2$ aperture, differing only in their vertical distributions to reflect specific clinical integration constraints: a baseline Planar Array (all sensors at $Z=20$\,mm), representing a standard under-bed setup; a Single-Split Array (outer columns at $Z=20$\,mm and inner columns at $Z=0$\,mm), which attempts to introduce parallax while remaining single-sided; and the proposed Staggered Split Array (outer columns at $Z=20$\,mm and inner columns at $Z=180$\,mm), which employs a dual-plate architecture to increase the observable information content.

\begin{figure}[t]
    \centering
    \includegraphics[width=\columnwidth]{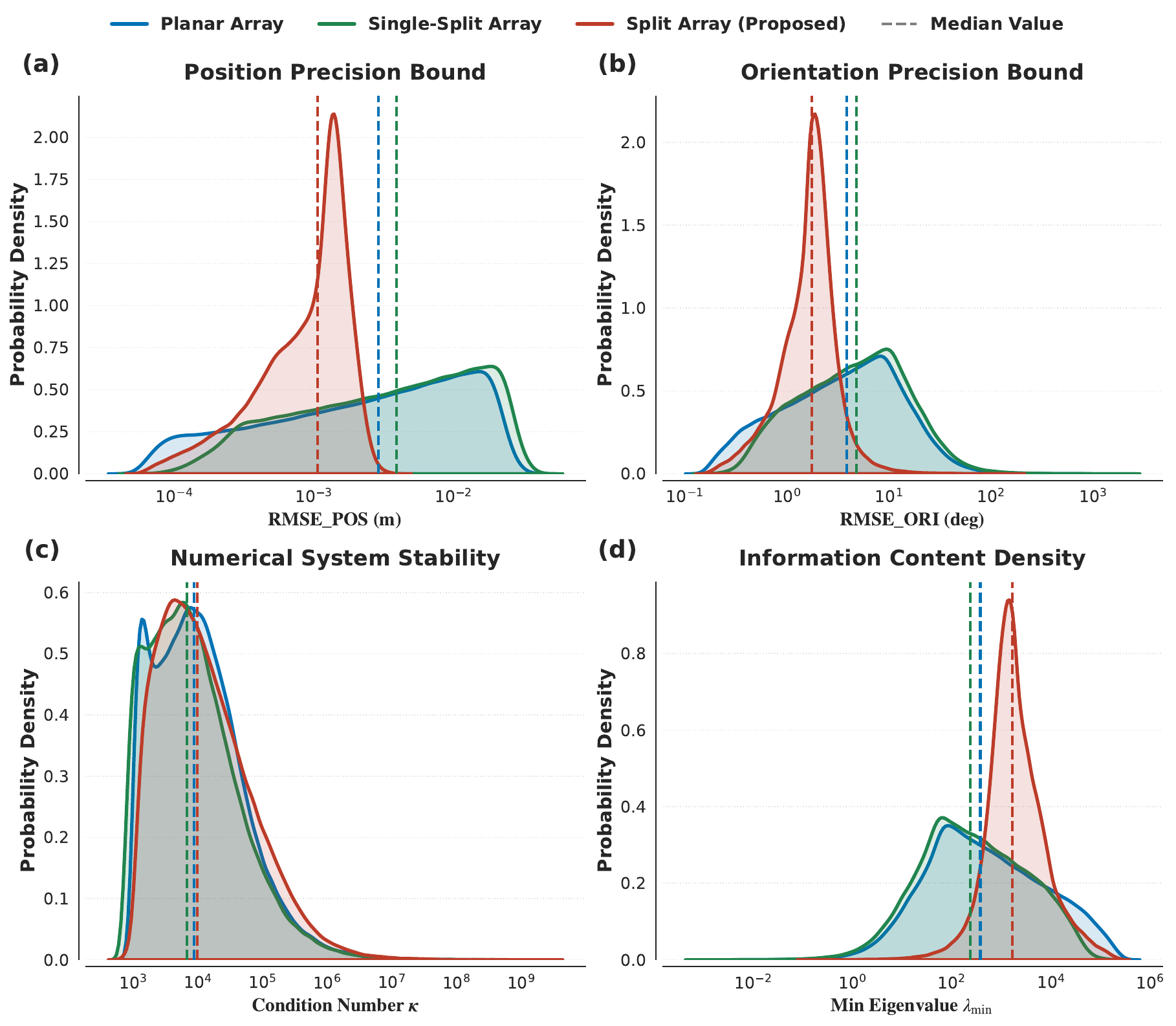}
    \caption{Comparative evaluation of Fisher Information-based observability metrics for three distinct sensor array configurations over a 200,000-sample validation set with $\sigma=10\,\mu\text{T}$ noise. 
    (a) and (b) show the theoretical lower bounds for position and orientation RMSE (CRLB); lower values indicate higher achievable precision. 
    (c) displays the condition number $\kappa$, representing numerical stability. 
    (d) illustrates the minimum eigenvalue $\lambda_{\min}$, quantifying the information density in the least observable direction. 
    While the ``Single-Split'' array (green) attempts to introduce depth parallax within a single-sided constraint, the proposed ``Staggered Split Array'' (red) significantly outperforms both benchmarks by providing stronger information density and improved precision bounds.
    }
    \label{fig:fim_comparison}
    \vspace{-6mm}
\end{figure}

The theoretical lower bounds on estimation error, visualized in Fig.~\ref{fig:fim_comparison}(a)--(b), reveal a decisive performance hierarchy. The proposed Staggered Split Array demonstrates superior observability, achieving a median position RMSE bound of 1.05\,mm and an orientation bound of 1.72$^{\circ}$. This corresponds to improvements of $2.7\times$ and $2.2\times$, respectively, over the Planar baseline (2.85\,mm and 3.76$^{\circ}$), confirming that extending the sensing aperture vertically effectively improves the information geometry. Crucially, however, the Single-Split Array yields an instructive negative result: despite introducing vertical offsets, its median position RMSE increases to 3.86\,mm, which is even worse than the Planar baseline. This counterintuitive finding highlights a fundamental physical constraint: achieving parallax by moving the inner sensors farther away (to $Z=0$\,mm) incurs a severe penalty due to the cubic decay of the magnetic field ($B \propto r^{-3}$). The resulting loss in signal-to-noise ratio (SNR) overwhelms the marginal gain in geometric observability, indicating that for single-sided arrays, sensor proximity to the target remains the dominant factor governing localization accuracy.

The underlying mechanism is further elucidated by the spectral properties of the FIM. As shown in Fig.~\ref{fig:fim_comparison}(d), the Staggered Array exhibits a median minimum eigenvalue ($\lambda_{\min}$) of $1.61 \times 10^{3}$, nearly an order of magnitude higher than those of the Planar ($3.62 \times 10^{2}$) and Single-Split ($2.27 \times 10^{2}$) arrays. Since $\lambda_{\min}$ quantifies the information density in the least observable direction, this confirms that the staggered topology effectively removes the dominant geometric blind spot. It is noteworthy that the condition number $\kappa$ of the proposed array ($9.60 \times 10^{3}$) is slightly higher than that of the Planar baseline ($8.48 \times 10^{3}$). While a larger $\kappa$ is often associated with reduced numerical stability, in this case it mainly reflects an expanded signal dynamic range: by placing the inner columns at $Z_{\text{TOP}}=180$\,mm, the array captures stronger near-field responses from the upper workspace, causing the dominant eigenvalue $\lambda_{\max}$ to increase disproportionately. Although this leads to greater anisotropy, the overall uncertainty volume ($\propto \det(\mathbf{F}^{-1})$) is still drastically reduced. Overall, these results show that the proposed Staggered Split Array breaks the geometry-induced information bottleneck and lowers the theoretical error floor to the millimeter level, whereas the Single-Split configuration provides no net benefit due to severe SNR attenuation; accordingly, the subsequent real-world experiments focus on the Proposed and Planar arrays only.

\subsection{Hardware Platform Setup}
\label{subsec:hardware}

\begin{figure}[t]
\centering
\includegraphics[width=0.9\columnwidth]{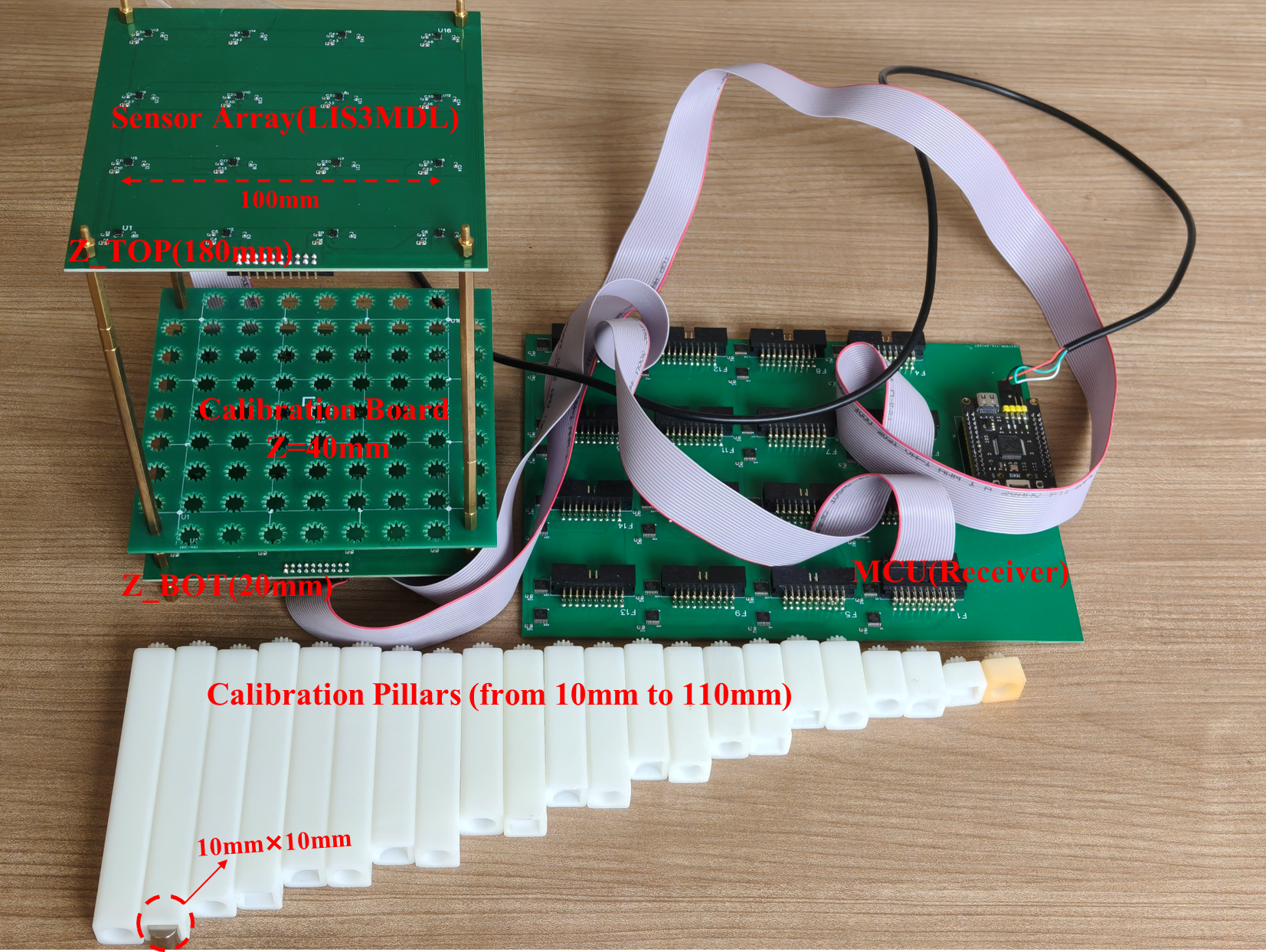}
\caption{The custom-designed hardware validation platform. The system features a dual-layer $4 \times 4$ sensor array positioned at $Z = 20$~mm and $Z = 180$~mm, a calibration board located at $Z = 40$~mm, a set of calibration pillars with varying heights, and an MCU-based acquisition unit. This configuration enables selective activation of 16-sensor subsets for a controlled comparison between planar and staggered spatial layouts under identical environmental conditions.}
\label{fig:hardware}
\vspace{-6mm}
\end{figure}

Guided by the observability analysis in Section~\ref{ssec:geom_observability_results}, we constructed a high-precision magnetic localization platform to acquire real-world validation data for both the Planar baseline and the proposed Staggered Split array. As illustrated in Fig.~\ref{fig:hardware}, the system comprises four primary subsystems: a custom-designed magnetometer array, a magnetic tracking target, a manual ground-truth positioning system, and a data acquisition unit.

\subsubsection{Sensing Array Topology and Configuration}

The core sensing component is a custom-designed magnetometer array based on LIS3MDL (STMicroelectronics) triaxial sensors. The platform adopts a dual-layer architecture consisting of two complete $4 \times 4$ sensor arrays (32 sensors in total) positioned at vertical heights of $Z = 20$~mm and $Z = 180$~mm, respectively. Within each layer, the sensors are uniformly distributed over a $100 \times 100$~mm planar region, spanning the horizontal coordinate range $x, y \in [-50, 50]$~mm.

The localization algorithm, however, does not use data from all 32 sensors simultaneously. Instead, the redundant hardware design allows us to selectively activate a subset of 16 sensors at a time while keeping the sensor count fixed and varying only the spatial topology. This enables a rigorous comparison between the proposed staggered-layer arrangement (using sensors from both layers) and a conventional planar arrangement (using sensors from a single layer), thereby isolating geometric topology as the sole variable affecting localization performance.

\subsubsection{Tracking Target and Physical Model}

The tracking target is an N35-grade cylindrical Neodymium (NdFeB) permanent magnet with dimensions of $\phi 10 \times 10$~mm. In the theoretical magnetic dipole model used for synthetic data generation, the magnet is characterized by the strength coefficient $B_T$, which is set to $7.9666 \times 10^{-2}$. This value ensures that the synthetic training data closely matches the physical properties of the N35 magnet used in the real-world experiments.

\subsubsection{Ground Truth and Data Processing}

To obtain rigorous ground truth, we employed a high-precision manual positioning system. Planar (XY) alignment was determined using a $100 \times 100$~mm calibration plate positioned at $Z = 40$~mm, featuring a $7 \times 7$ grid of positioning holes spaced 15~mm apart, while vertical (Z-axis) displacement was controlled using a set of custom-fabricated resin calibration pillars with precisely controlled heights.

Data acquisition is orchestrated by an STM32F103RCT6 microcontroller (MCU), which synchronously samples the active sensors at 100~Hz. The collected data is transmitted to a workstation running Ubuntu 22.04, equipped with an Intel(R) Xeon(R) Platinum 8488C CPU and an NVIDIA GeForce RTX PRO 6000 GPU for subsequent data processing and model training. No explicit post-fabrication calibration was performed for the sensor coordinates or sensitivities. Instead, the system directly uses the nominal array geometry and manufacturer-specified sensor characteristics. Accordingly, all real-world experiments are conducted in a calibration-free setting.

\subsection{Datasets}
\label{subsec:dataset}

To enable robust Sim-to-Real learning without manual calibration, we constructed two distinct datasets: a large-scale synthetic dataset for training and a real-world dataset used exclusively for evaluation. Both datasets share the same coordinate system and sensor geometry defined in Section~\ref{subsec:hardware}.

We generated the training data using a physics-based simulation based on Eq.~(\ref{eq:B_vector}). To reduce the Sim-to-Real gap, we adopted a hardware-aware strategy by clipping the simulated magnetic flux density at the sensor saturation threshold ($\pm 1900\,\mu\mathrm{T}$). This enables the network to learn how to handle saturated signals commonly encountered in near-field tracking. We used Latin Hypercube Sampling (LHS)~\cite{Garg03042017} to generate uniformly distributed pose samples across the workspace ($x, y \in [-50, 50]$~mm, $z \in [45, 155]$~mm). Using this sampling strategy, we created two datasets with $1 \times 10^7$ samples each: the \textit{Staggered Split Dataset}, constructed from the dual-layer coordinates ($Z \in \{20, 180\}$~mm), and the \textit{Planar Dataset}, constructed from the lower layer only.

\begin{figure}[t]
\centering
\includegraphics[width=\columnwidth]{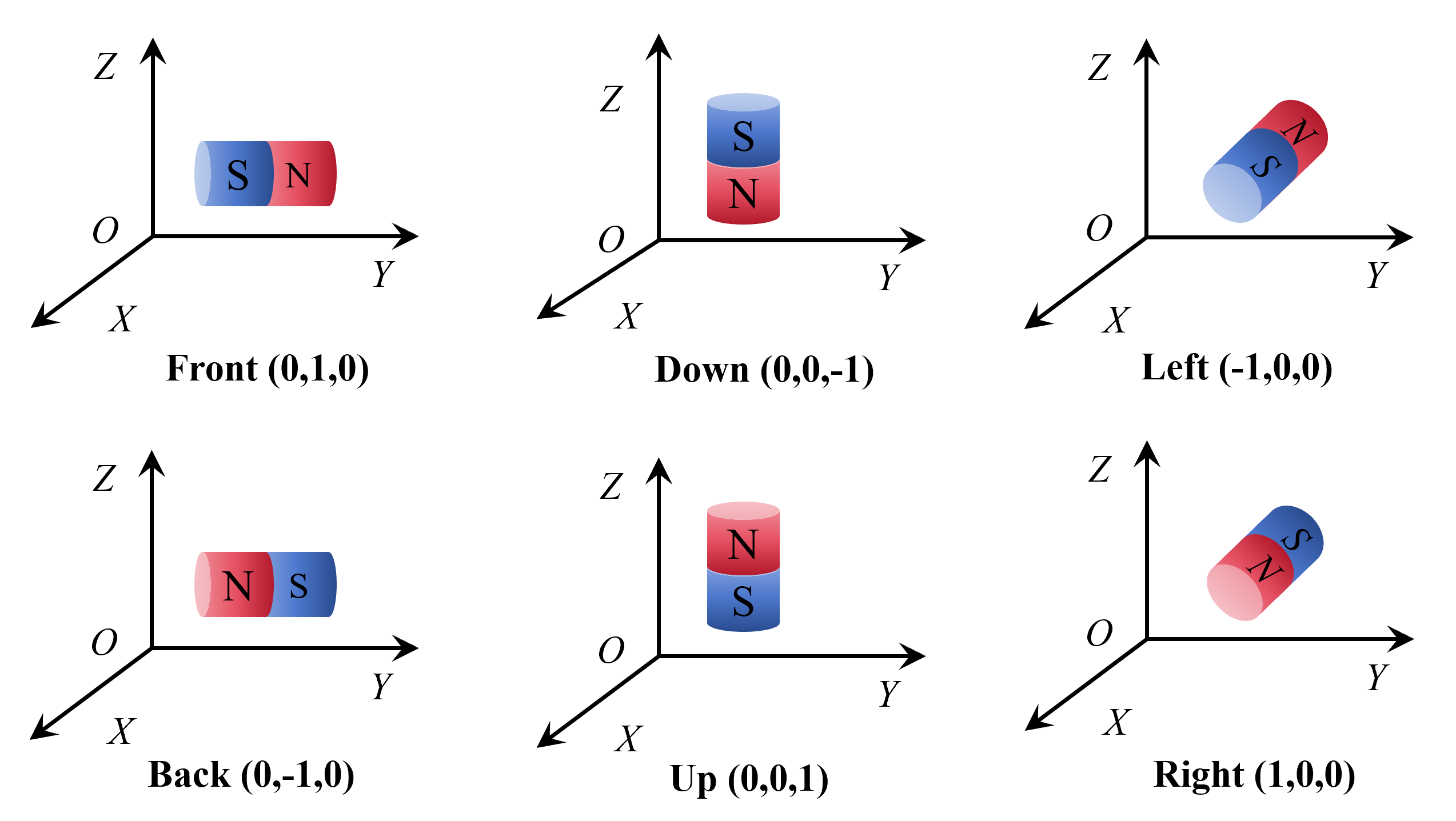}
\caption{The six magnet orientations used for data acquisition. The corresponding unit orientation vectors are Front $(0,1,0)$, Back $(0,-1,0)$, Up $(0,0,1)$, Down $(0,0,-1)$, Left $(-1,0,0)$, and Right $(1,0,0)$.}
\label{fig_7}
\vspace{-6mm}
\end{figure}

For physical validation, we collected a real-world dataset using the high-precision platform. We adopted a grid sampling strategy by placing the magnet at 49 uniformly distributed positions on the XY plane (a $7 \times 7$ grid with 15~mm spacing). At each position, data were recorded at 11 discrete heights from $Z = 50$~mm to $Z = 150$~mm in steps of 10~mm. We also rotated the magnet to six fixed orientations (Fig.~\ref{fig_7}) at each point, yielding a total of 3,234 real-world samples.

To ensure a fair comparison under identical environmental conditions, we extracted sensor readings for both topologies from the same 32-sensor hardware recordings. For the Staggered configuration, we selected data from the outer columns of the lower layer and the inner columns of the upper layer. For the Planar baseline, we used only the lower layer data. This real-world dataset is fully separated from training and used only for testing.

\subsection{Training and Evaluation Criteria}
\label{subsec:train_eval}

\subsubsection{Training and Implementation Details}

The proposed Phy-GAANet was implemented in PyTorch 2.9.1 with CUDA 12.8 and cuDNN 9.1 acceleration. All training and evaluation experiments were conducted on a high-performance workstation equipped with an Intel(R) Xeon(R) Platinum 8488C CPU and an NVIDIA RTX PRO 6000 GPU.

We trained the network using the AdamW optimizer for 150 epochs with a batch size of 2,048. The initial learning rate was set to $1 \times 10^{-3}$ with a weight decay of $1 \times 10^{-5}$, and a Cosine Annealing scheduler was employed to gradually reduce the learning rate throughout training for improved convergence stability. To enhance robustness against real-world sensor fluctuations, Gaussian noise with a standard deviation of 2\% of the signal magnitude ($\sigma = 0.02$) was injected into the input data online during training. The checkpoint with the lowest validation position error was selected for final testing.

Notably, the entire training pipeline was based exclusively on synthetic data generated from the theoretical physics model, and the validation set was also drawn from the synthetic domain. This fully synthetic training paradigm eliminates the need for labor-intensive real-world data collection during training. Details of the dataset generation procedure are provided in Section~\ref{subsec:dataset}.

\subsubsection{Evaluation Metrics}

To comprehensively assess localization performance, we use two primary metrics: Position Error ($E_{pos}$) and Orientation Error ($E_{ori}$).

\textbf{Position Error ($E_{pos}$):} Defined as the Euclidean distance between the estimated position $\hat{\mathbf{p}} \in \mathbb{R}^{3}$ and the ground truth $\mathbf{p} \in \mathbb{R}^{3}$:
\begin{equation}
    E_{pos} = \|\hat{\mathbf{p}} - \mathbf{p}\|_2 \times 1000 \quad \text{(mm)}
\end{equation}

\textbf{Orientation Error ($E_{ori}$):} Defined as the angular deviation between the estimated unit orientation vector $\hat{\mathbf{n}}$ and the ground truth $\mathbf{n}$:
\begin{equation}
    E_{ori} = \arccos\left(\frac{\hat{\mathbf{n}} \cdot \mathbf{n}}{\|\hat{\mathbf{n}}\|_2 \|\mathbf{n}\|_2}\right)\times\frac{180}{\pi} \quad \text{(deg)}
\end{equation}
where $\cdot$ denotes the dot product. Although the model operates on unit vectors (implying $\|\hat{\mathbf{n}}\|_2 \approx 1$ and $\|\mathbf{n}\|_2 = 1$), the denominator is retained to ensure numerical robustness.

\subsection{Evaluation of Localization Performance}
\label{subsec:ELA}

\begin{table}[t]
\centering
\caption{Global Localization Performance Comparison}
\label{tab:global_comparison}
\setlength{\tabcolsep}{1.5pt}
\resizebox{\columnwidth}{!}{%
\begin{tabular}{l|ccc|ccc|c}
\hline
\multirow{2}{*}{\textbf{Method}} & \multicolumn{3}{c|}{\textbf{Pos. Error (mm)}} & \multicolumn{3}{c|}{\textbf{Ori. Error ($^\circ$)}} & \textbf{Time} \\ \cline{2-8} 
 & \textbf{Mean$\pm$STD} & \textbf{RMSE} & \textbf{Max} & \textbf{Mean$\pm$STD} & \textbf{RMSE} & \textbf{Max} & \textbf{(ms)} \\ \hline
LM (Planar) & 5.12 $\pm$ 4.62 & 6.90 & 50.90 & 5.49 $\pm$ 7.10 & 8.98 & 132.41 & 2.31 \\
LM (Split) & 2.95 $\pm$ 4.67 & 5.52 & 75.08 & 4.79 $\pm$ 10.28 & 11.34 & 162.80 & 2.12 \\
Phy-GAANet (Planar) & 3.10 $\pm$ 2.97 & 4.30 & 24.29 & 4.55 $\pm$ 4.18 & 6.18 & 29.87 & 3.38 \\
\textbf{Phy-GAANet (Split)} & \textbf{1.84 $\pm$ 1.34} & \textbf{2.27} & \textbf{10.10} & \textbf{3.18 $\pm$ 2.02} & \textbf{3.84} & \textbf{17.94} & 3.64 \\ \hline
\end{tabular}%
}
\end{table}

\begin{figure}[t]
    \centering
    \includegraphics[width=\columnwidth]{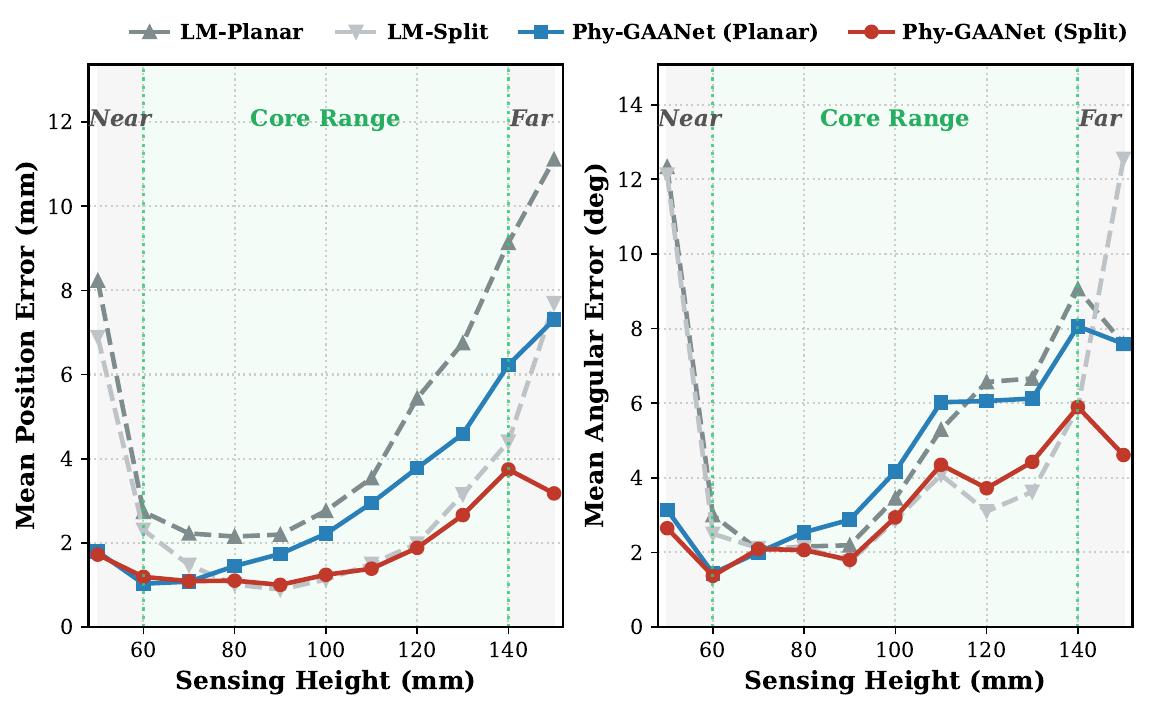}
    \vspace{-6mm}
    \caption{Layer-wise localization error distribution along the $Z$ axis. The shaded regions indicate the boundary-sensitive regimes around $Z = 50$~mm and $Z = 150$~mm. The proposed Phy-GAANet with the Staggered Split Array (red curve) maintains consistently lower position and orientation errors across the workspace, whereas the LM-based optimization exhibits pronounced degradation near the workspace boundaries.}
    \label{fig:comparison_results}
    \vspace{-6mm}
\end{figure}

To thoroughly validate the proposed framework, we benchmark Phy-GAANet against the classical Levenberg--Marquardt (LM) iterative optimization algorithm on the real-world dataset. For a rigorous comparison, the LM solver was initialized with a fixed perturbation from the ground truth (position: $+20$~mm; orientation vector components: $+0.3$), consistent with the protocol in prior work~\cite{xie2025theoretical}. This comparison allows us to decouple the contributions of algorithmic learning and geometric optimization. As summarized in Table~\ref{tab:global_comparison}, the results demonstrate a clear advantage of our approach across multiple performance dimensions.

The global performance metrics reveal a clear hierarchy driven by both algorithmic and geometric improvements. Algorithmically, Phy-GAANet consistently outperforms the classical LM solver regardless of the sensor topology. On the conventional Planar array, Phy-GAANet reduces the position RMSE from 6.90~mm (LM) to 4.30~mm. Similarly, on the Split array, it further lowers the error from 5.52~mm (LM) to 2.27~mm, demonstrating that the data-driven model handles real-world noise more effectively than iterative optimization. Geometrically, the Staggered Split topology provides intrinsic advantages over the Planar baseline, independent of the estimator used. Both LM and Phy-GAANet achieve substantially lower errors when transitioning from Planar to Split configurations. Consequently, by combining the best-performing algorithm with the best-performing geometry, the Phy-GAANet (Split) configuration achieves the best overall performance across all metrics (position RMSE: 2.27~mm; orientation RMSE: 3.84$^\circ$).

Crucially, the maximum-error statistics reveal a strong robustness advantage of the proposed learning-based approach. While the LM algorithm benefits on average from the Split geometry, its dependence on initialization makes it prone to catastrophic divergence in ill-conditioned regions. This results in unacceptable worst-case outliers, with maximum position errors reaching 75.08~mm, essentially indicating tracking failure. In contrast, Phy-GAANet behaves as a much more stable estimator. By embedding physical constraints, it effectively suppresses extreme outliers, limiting the maximum position error to 10.10~mm even in the most challenging scenarios. This robustness is particularly important for clinical applications, where reliability matters as much as average accuracy.

The system's robustness is further illustrated by the boundary regions of the workspace ($Z = 50$~mm and $Z = 150$~mm), as shown in Fig.~\ref{fig:comparison_results}. At the lower boundary ($Z = 50$~mm), the magnet is only 30~mm above the bottom sensors, where high magnetic flux density frequently causes sensor saturation and deviation from the point-dipole assumption. In this regime, the LM algorithm exhibits pronounced error spikes (Planar mean: 8.23~mm), whereas Phy-GAANet maintains a low mean position error of 1.72~mm, indicating that the network effectively compensates for near-field distortions and saturation artifacts. At the upper boundary ($Z = 150$~mm), the Planar topology suffers from severe far-field signal attenuation. In contrast, under the Split topology, the magnet enters a new near-field regime relative to the top sensors ($Z_{\text{TOP}} = 180$~mm). Leveraging this symmetric proximity, Phy-GAANet reduces the mean error to 3.18~mm, effectively bridging the accuracy gap typically observed at the workspace boundary.

Regarding computational efficiency, although the proposed network incurs a marginal increase in inference time (3.64~ms) compared with the LM algorithm (2.12~ms), this trade-off is negligible in practice. The inference speed corresponds to a theoretical update rate exceeding 270~Hz, which is more than sufficient for real-time medical guidance. Furthermore, unlike the iterative LM solver, which suffers from variable convergence times, Phy-GAANet provides deterministic constant-time inference, thereby ensuring temporal stability for real-time control loops.

\subsection{Comparison With Existing Approaches}
\label{subsec:CWEA}

It is important to note that permanent-magnet localization is a highly specialized problem. Beyond algorithmic design, localization performance is strongly influenced by hardware variations, including magnet dimensions, magnetization intensity, array geometry, and manufacturing tolerances. Consequently, directly comparing error metrics across different studies without reproducing the underlying hardware setup is of limited significance. To provide a fairer comparison, we compiled Table~\ref{tab:comparison} under a standardized benchmark in which the hardware platform was unified, state-of-the-art algorithms were re-implemented, and identical training settings were applied without prior calibration. This benchmark is intended to evaluate the relative merits of the proposed architecture under controlled conditions.

\begin{table}[t]
\centering
\caption{Comparison with State-of-the-Art Approaches}
\label{tab:comparison}
\setlength{\tabcolsep}{1.2pt} 
\resizebox{\columnwidth}{!}{%
\begin{tabular}{l|ccc|ccc|c}
\hline
\multirow{2}{*}{\textbf{Method}} & \multicolumn{3}{c|}{\textbf{Pos. Error (mm)}} & \multicolumn{3}{c|}{\textbf{Ori. Error ($^\circ$)}} & \textbf{Time} \\ \cline{2-8} 
 & \textbf{Mean} & \textbf{RMSE} & \textbf{Max} & \textbf{Mean} & \textbf{RMSE} & \textbf{Max} & \textbf{(ms)} \\ \hline
FCN-Indep~\cite{11151691} & 3.03 & 3.91 & 26.41 & 4.10 & 5.83 & 67.73 & \textbf{0.87} \\
PIRNet~\cite{11246085} & 2.14 & 2.86 & 20.98 & 3.54 & 4.63 & 26.32 & 1.03 \\
AMagPosenet~\cite{su2023amagposenet} & 1.98 & 2.60 & 16.91 & 3.94 & 5.37 & 36.71 & 1.09 \\
MobilePoseNet~\cite{xie2025theoretical} & 2.28 & 2.85 & 14.88 & 3.47 & 4.08 & 14.88 & 2.54 \\
ResNet-18~\cite{guo2022improved} & 1.91 & 2.49 & 24.02 & 3.52 & 4.49 & 25.72 & 1.94 \\ \hline
\textbf{Phy-GAANet} & \textbf{1.84} & \textbf{2.27} & \textbf{10.10} & \textbf{3.18} & \textbf{3.84} & \textbf{17.94} & 3.64 \\ \hline
\end{tabular}%
}
\vspace{-4mm}
\end{table}

\begin{figure}[t]
    \centering
    \includegraphics[width=\columnwidth]{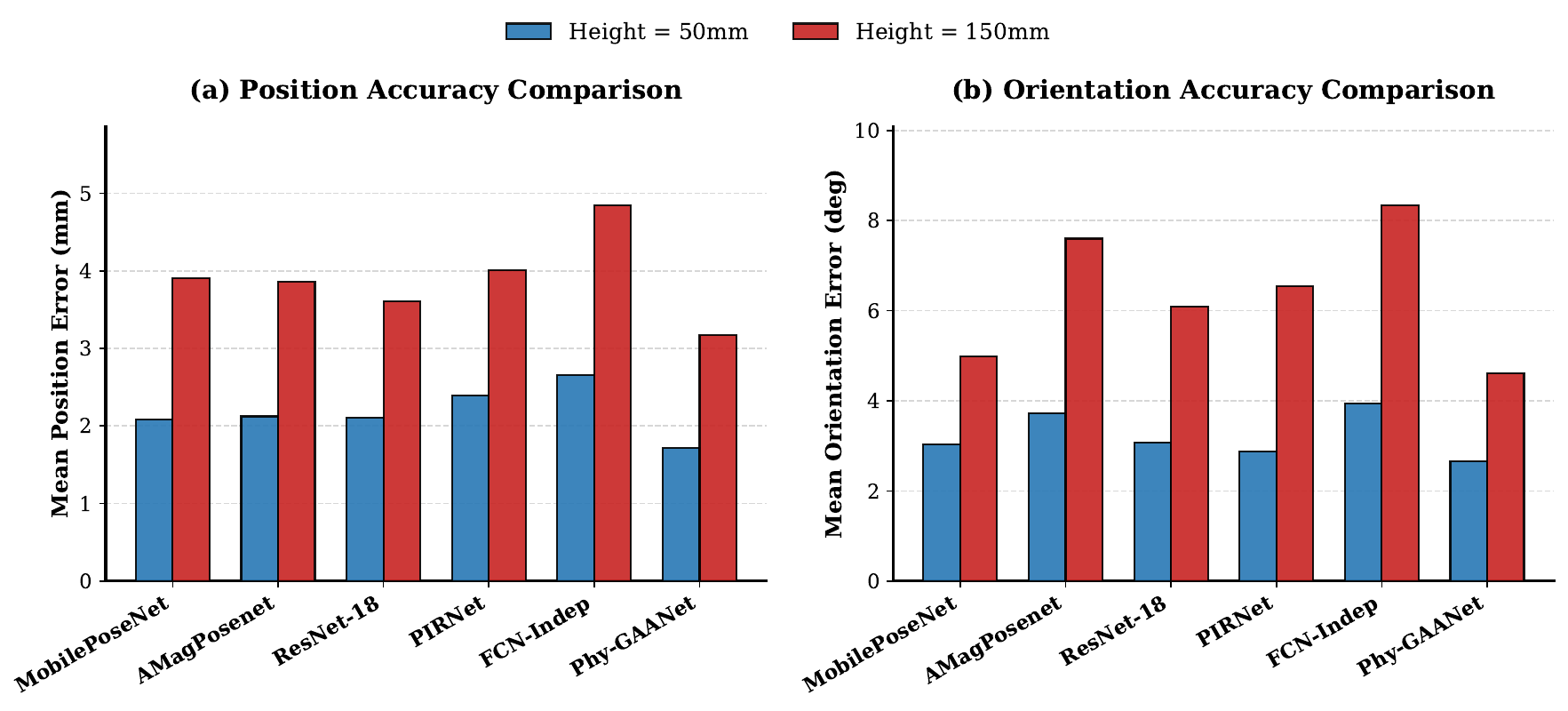} 
    \caption{Performance comparison at the critical boundary heights $Z = 50$~mm and $Z = 150$~mm. Under the Staggered Split Array topology, both heights correspond to near-field regimes where partial sensor saturation occurs and the dipole model begins to break down. Panels (a) and (b) compare the mean position and orientation errors, respectively, across representative baseline methods and the proposed Phy-GAANet.}
    \label{fig:near_field_comparison}
    \vspace{-4mm}
\end{figure}

The quantitative results in Table~\ref{tab:comparison} show that \textit{Phy-GAANet} achieves state-of-the-art performance. Our method attains the lowest mean position error among all competing architectures. More importantly, its most significant advantage lies in stability, as reflected by the maximum-error metric. While strong backbones such as ResNet-18 achieve competitive mean errors, they still suffer from substantial outliers (Max: 24.02~mm) due to the lack of explicit physical constraints. In contrast, Phy-GAANet behaves as a much more stable estimator. By explicitly encoding geometric priors, it limits the maximum position error to 10.10~mm, corresponding to a 58\% reduction relative to ResNet-18. This indicates that the proposed architecture effectively suppresses catastrophic divergence in challenging signal regimes.

To further assess robustness under extreme physical constraints, we analyzed performance at the two workspace boundaries ($Z = 50$~mm and $Z = 150$~mm). In the Staggered Split configuration, both correspond to near-field conditions in which the magnet is only 30~mm from the nearest sensor layer. These cases are particularly challenging because they combine partial sensor saturation with deviation from the point-dipole assumption. As shown in Fig.~\ref{fig:near_field_comparison}, existing methods degrade noticeably in this regime; for example, MobilePoseNet exhibits position errors of nearly 4.0~mm at $Z = 150$~mm, indicating limited ability to separate informative high-gradient measurements from saturation artifacts. By contrast, Phy-GAANet maintains strong robustness through its GAA module, which adaptively emphasizes features from the non-saturated layer, thereby achieving lower mean position and orientation errors at the workspace boundaries.

\subsection{Ablation Study}
\label{subsec:AS}

To investigate the contributions of the proposed components within \textit{Phy-GAANet}, we conducted a comprehensive ablation study on the same real dataset. We systematically removed or replaced key modules to evaluate their effects on localization accuracy and robustness. The evaluated variants are defined as follows: (1) \textbf{w/o PIF}: removes the Physics-Informed Features (logarithmic magnitude and saturation mask), relying solely on raw magnetic flux density; (2) \textbf{w/o Attn}: removes all attention mechanisms from the backbone; (3) \textbf{w/ SE-Block}: replaces the proposed physics-aware GAA module with the standard SE block~\cite{hu2018squeeze} to benchmark against generic channel attention; and (4) \textbf{GAA Variants}: investigates the internal mechanism of the GAA module by selectively disabling its \textit{Group} (axis-wise), \textit{Gradient} (local spatial), or \textit{Channel} (global) branches.

\begin{table}[t]
\centering
\caption{Ablation Study on Model Components}
\label{tab:ablation_results}
\setlength{\tabcolsep}{1.2pt}
\resizebox{\columnwidth}{!}{%
\begin{tabular}{l|ccc|ccc}
\hline
\multirow{2}{*}{\textbf{Configuration}} & \multicolumn{3}{c|}{\textbf{Pos. Error (mm)}} & \multicolumn{3}{c}{\textbf{Ori. Error ($^\circ$)}} \\ \cline{2-7}
 & \textbf{Mean $\pm$ STD} & \textbf{RMSE} & \textbf{Max} & \textbf{Mean $\pm$ STD} & \textbf{RMSE} & \textbf{Max} \\ \hline
Phy-GAANet (Full) & 1.84 $\pm$ 1.34 & 2.27 & 10.10 & 3.18 $\pm$ 2.02 & 3.84 & 17.94 \\ \hline
\textit{w/o} PIF & 1.85 $\pm$ 1.33 & 2.28 & 10.92 & 3.45 $\pm$ 2.54 & 4.28 & 24.94 \\ \hline
\textit{w/o} Attn & 1.97 $\pm$ 1.38 & 2.41 & 10.33 & 3.28 $\pm$ 1.96 & 3.82 & 17.36 \\
\textit{w/} SE-Block & 2.01 $\pm$ 1.47 & 2.49 & 11.13 & 3.37 $\pm$ 2.21 & 4.03 & 22.26 \\ \hline
\textit{GAA w/o} Group & 2.02 $\pm$ 1.48 & 2.50 & 11.51 & 3.33 $\pm$ 2.06 & 3.91 & 18.87 \\
\textit{GAA w/o} Gradient & 1.95 $\pm$ 1.39 & 2.40 & 10.01 & 3.24 $\pm$ 1.95 & 3.78 & 16.44 \\
\textit{GAA w/o} Channel & 1.97 $\pm$ 1.43 & 2.44 & 11.28 & 3.25 $\pm$ 1.99 & 3.81 & 20.05 \\ \hline
\end{tabular}%
}
\vspace{-6mm}
\end{table}

The quantitative results are summarized in Table~\ref{tab:ablation_results}. First, the Physics-Informed Features (PIF) play an important role in orientation stability. While removing PIF (\textit{w/o} PIF) causes only a marginal change in position accuracy, it leads to noticeably degraded orientation robustness: the angular RMSE increases to 4.28$^\circ$, and the maximum orientation error reaches 24.94$^\circ$, compared with 17.94$^\circ$ for the full model. This suggests that the explicit inclusion of saturation masks and logarithmic scaling is particularly beneficial for suppressing extreme outliers in rotation estimation.

Second, the comparison of attention mechanisms highlights the importance of the proposed GAA module. The full Phy-GAANet achieves the best overall balance between position accuracy and robustness, with the lowest mean position error (1.84~mm) and the smallest position uncertainty ($\pm 1.34$~mm). In contrast, replacing GAA with a standard SE block (\textit{w/} SE-Block) degrades the mean position error to 2.01~mm and increases the maximum position error to 11.13~mm, even performing worse than the backbone without attention. This indicates that generic channel attention is insufficient to capture the inherent information of the magnetic field, whereas GAA is better aligned with the underlying sensing physics.

Finally, the internal decomposition of GAA clarifies the contribution of each branch. Disabling the Group branch (\textit{GAA w/o} Group) yields the worst position performance among all ablation variants (RMSE 2.50~mm, Max 11.51~mm), showing that modeling the correlation within sensor triads ($B_x, B_y, B_z$) is crucial for robust localization. Removing the Channel branch also degrades stability, particularly in the maximum orientation error. Although removing the Gradient branch leads to slightly lower max-error statistics in a few cases, the full model remains the most balanced configuration overall.

\subsection{Discussion}
\label{Sec:DIS}

In practical deployment, rigid sensor carriers are generally preferred to flexible or bendable structures because they are easier to fabricate, calibrate, and maintain. Within this rigid external-placement setting, possible sensor layouts range from a single planar array to vertically separated planar layers, and even to a cubic shell that fully surrounds the workspace. Although cubic arrays can generally provide superior sensing performance, they are not a practical primary hardware solution for our system, because external magnetic actuation typically requires access from multiple directions around the workspace; installing sensors on all six exterior faces would directly restrict the placement and operation of the external magnetic driving apparatus. Accordingly, this study focused on low-complexity geometric modifications along a single structural degree of freedom, namely the vertical redistribution of sensor layers, and confirmed that the staggered split arrangement substantially outperforms a planar baseline.

\begin{figure}[t]
    \centering
    \includegraphics[width=\columnwidth]{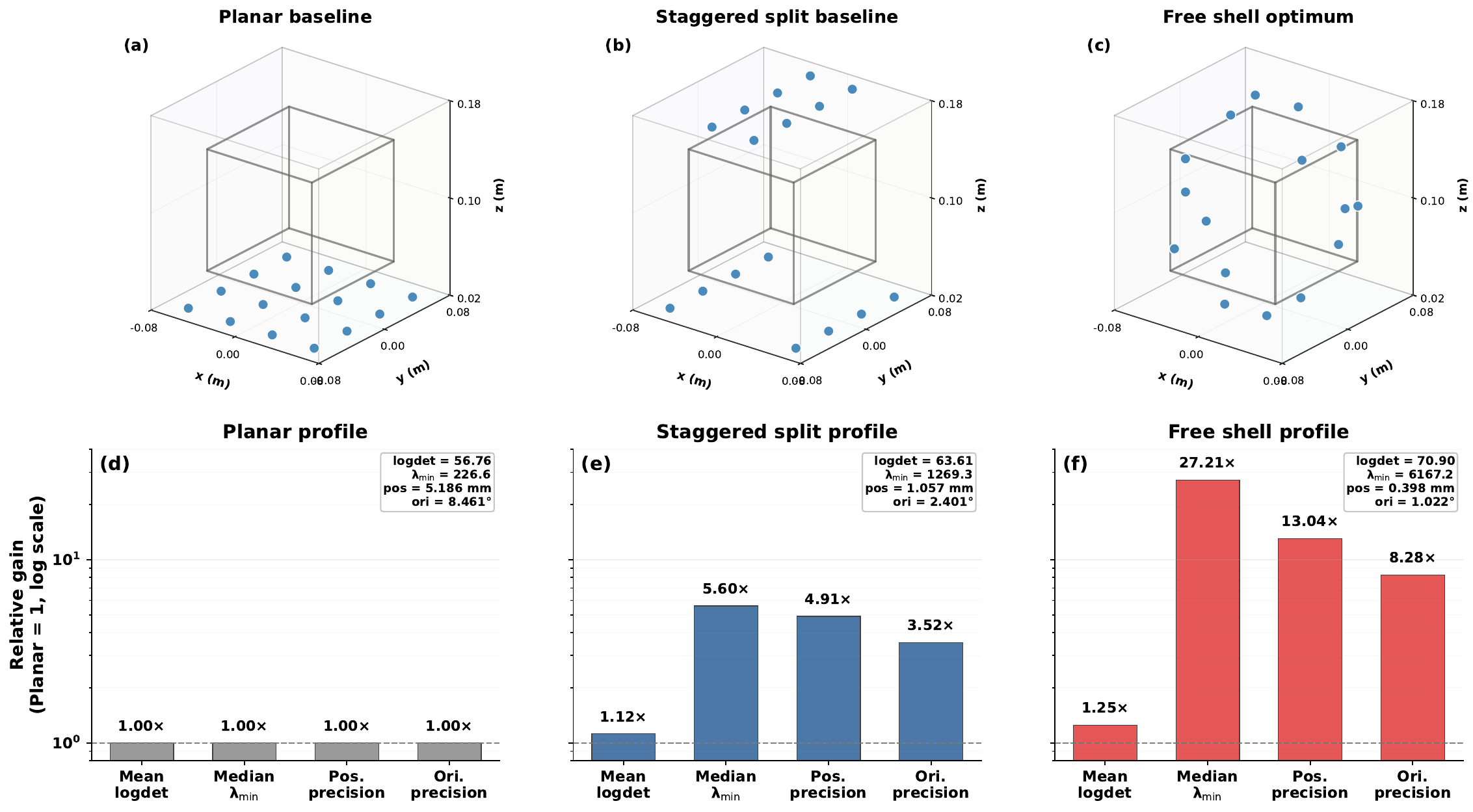}
    \caption{Top row: comparison of three rigid external sensor layouts, including the planar baseline, the staggered split baseline, and the shell-constrained free optimum. Bottom row: corresponding normalized performance profiles under a shared logarithmic y-axis, where larger values indicate better performance. The four metrics are mean log-determinant, median minimum eigenvalue, position precision, and orientation precision, with the latter two defined relative to the planar baseline through the inverse CRLB.}
    \label{fig:free_shell_layouts}
\vspace{-6mm}
\end{figure}

Nevertheless, the cubic shell-constrained optimum remains valuable as a theoretical reference, because it reveals how much geometric headroom still exists under rigid external placement. To examine this, we extended the same 5-DOF Fisher-information framework and performed a free-placement optimization over the six outer faces of a cubic shell with side length 0.16\,m, using a two-stage strategy consisting of greedy initialization followed by local continuous refinement. As shown in Fig.~\ref{fig:free_shell_layouts}, the resulting optimum is not a uniform distribution, but a structured multi-face arrangement, indicating that informative sensor placement is governed by observability complementarity across different viewing directions. Quantitatively, the staggered split baseline already achieves a substantial improvement over the planar layout, showing that even a limited vertical redistribution of sensors can recover much of the information gain enabled by three-dimensional sensing diversity. Building on this, the shell-constrained optimum further improves the geometry by increasing the overall information content, strengthening the weakest observable mode, and reducing the mean position and orientation CRLBs from 1.057\,mm and 2.401$^\circ$ to 0.398\,mm and 1.022$^\circ$, respectively. These results indicate that the proposed theoretical framework is capable of identifying substantially more accurate geometric configurations under higher degrees of placement freedom, highlighting its potential as a tool for sensor layout design in applications requiring high localization precision. Such configurations, however, should be interpreted as constrained theoretical references rather than practical hardware targets. In contrast, the proposed staggered split array already provides a substantial geometric advantage over the planar baseline while preserving workspace openness and maintaining compatibility with external magnetic actuation.

\section{Conclusion}
\label{Sec:CONC}

This work presents a unified framework for high-precision magnetic localization that combines information-theoretic geometry optimization with physics-aware deep learning. FIM-based analysis shows that the proposed staggered split-array topology provides a substantially stronger observability foundation than a planar baseline. Built on this geometry, Phy-GAANet bridges the Sim-to-Real gap through physics-informed input design and Geometry-Aware Attention, enabling robust calibration-free localization from raw sensor measurements. Real-world experiments demonstrate state-of-the-art performance, achieving 1.84\,mm position error and 3.18$^{\circ}$ orientation error at over 270\,Hz, while maintaining robustness in challenging near-field regions. Future work will improve interference robustness and integrate the framework with external magnetic actuation for clinical deployment.
% References

\bibliographystyle{Bibliography/IEEEtranTIE}
\bibliography{ref11} %IEEEabrv instead of IEEEfull

\end{document}